\documentclass[11pt]{article}

\usepackage[utf8]{inputenc}
\usepackage[T1]{fontenc}
\usepackage{lmodern}
\usepackage[margin=1in]{geometry}
\usepackage{amsmath,amssymb}
\usepackage{booktabs,longtable,array,calc,tabularx,float}
\newcolumntype{Y}{>{\raggedright\arraybackslash}X}
\newcolumntype{Z}{>{\centering\arraybackslash}X}
\usepackage[round,authoryear]{natbib}
\usepackage{microtype}
\usepackage{etoolbox}
\usepackage{xcolor}
\usepackage[hidelinks]{hyperref}
\usepackage{url}
\usepackage{caption}
\providecommand{\tightlist}{\setlength{\itemsep}{0pt}\setlength{\parskip}{0pt}}
\AtBeginEnvironment{longtable}{\small\setlength{\tabcolsep}{4pt}}
\AtBeginEnvironment{verbatim}{\small}
\usepackage{fvextra}
\DefineVerbatimEnvironment{verbatim}{Verbatim}{breaklines=true,fontsize=\small}

\title{Copies Are Not Confirmations:\\How News Duplication Fools LLM Judges and Continual Learners}

\author{
  Abdelkader Metales\\
  caaise GmbH\\
  \texttt{metales@caaise.com}
  \and
  Miriam Grabenheinrich\\
  caaise GmbH\\
  \texttt{grabenheinrich@caaise.com}
  \and
  Aicha Metales\\
  Djilali Liabes University of Sidi Bel Abbes\\
  \texttt{aicha.metales@univ-sba.dz}
}

\date{}

\begin{document}
\maketitle

\begin{abstract}
A language model that keeps learning from the news sees the same claim many times, but many reports are copies of one source: wire stories, content shared within a media group, syndication and press releases. We ask whether LLM judges and continual learners treat such copies as independent confirmations. In 200,000 English articles from CC-News, a near-duplicate detector identifies copy links to another outlet for 18.9\% of the articles, a detector-based lower bound. In a controlled experiment with claims about invented people, three identical reprints instead of one report raised the lies accepted by Llama 3.2 from 0.06 to 0.71 and by phi4 from 0.21 to 0.90; Gemini 2.5 Flash already believed a single report, and Gemini 2.5 Pro followed provenance information most consistently. In a simulated continual-learning stream, showing the judge one report per detected copy chain cut the copied lies accepted by 0.10-0.24 across four judges, at a mean cost of at most 4.2 points of accepted truths. Text similarity links none of the copies reworded by three LLMs. An entity-novelty test, the share of a report's named entities absent from an earlier look-alike, separates these rewrites from 100 author-labelled independent reports of the same event (AUC 0.98-0.99), statistically indistinguishable from TF-IDF and the best tested embedding model, but recognises only 17\% of natural same-source derivatives. With both detectors run on 15,084 real articles carrying inserted claims, one report per chain cut the lies accepted by phi4 from 0.57-0.67 to 0.09 (copies), 0.14 (rewrites) and 0.34 (rewrites with invented details), and by Llama 3.2 from 0.56-0.62 to 0.07, 0.09 and 0.25, at a cost of 2.4 and 10 points of accepted truths. Counting reports or chains without a judge is unreliable, and colluding sources that do not copy each other remain an open problem.

\end{abstract}

\section{Introduction}\label{sec:intro}

Models that update themselves from incoming text must decide which claims to believe. In our earlier work, a continual learner was safe only to the extent that a judge (a verifier that approves each claim) was accurate, and a learner that selected claims by surprise absorbed nearly every lie \citep{metales2026surprise}. News adds a further problem. The same claim reaches a reader many times, but often through copies: a wire story reprinted by dozens of papers, articles shared across the titles of one owner, syndicated content and press releases. Journalists treat such reprints as one source, not many. A learner or judge that treats every publication as independent evidence may therefore be swayed by claims republished from a single source.

Repetition is known to shift what language models believe: repeated counterfactual claims in continued pre-training flip a large share of beliefs \citep{churina2025layer}, and repeated low-credibility information in the prompt reverses models' source preferences \citep{schuster2026whose}. Neither line of work separates copies from independent sources, and neither tests a defence inside a continual learner.

\textbf{Terms.} A \emph{report} is one article stating a claim. A \emph{copy} repeats an earlier report: a verbatim \emph{reprint}, an \emph{edited copy} (credit or lines removed) or a \emph{rewrite} (same content in new wording, here produced by an LLM). A \emph{chain} is a group of reports our detector links as copies; it is a duplicate cluster, not a verified provenance lineage. \emph{Independent reports} gather their information separately; \emph{collusion} means distinct outlets that coordinate on the same false claim without copying one another: they look independent but are not. We keep four properties apart: textual similarity, which our detectors measure; information dependence, whether one report derives from another, which they estimate; source independence, separate information gathering; and corroboration, whether the claim is true. This paper establishes neither of the last two.

We ask four questions:

\begin{itemize}
\tightlist
\item
  \textbf{RQ1} How common are copies in real news, and what kinds of dependence do they reflect? (Section~\ref{sec:copying})
\item
  \textbf{RQ2} Do LLM judges treat copies as independent confirmations, and does telling them otherwise help? (Section~\ref{sec:judges})
\item
  \textbf{RQ3} In a continual learner fed by a stream of reports, do rules that count reports or independent chains keep lies out? (Section~\ref{sec:stream})
\item
  \textbf{RQ4} Does removing copies before the judge sees the evidence reduce the lies learned, and against which attacks? (Sections~\ref{sec:stream} and~\ref{sec:real})
\end{itemize}

The paper studies whether repeated publications are treated as independent evidence by LLM judges and learning pipelines, and whether filtering by dependence changes that behaviour. Our contributions fall into four groups. (1) \emph{Measurement}: cross-outlet copying in 200,000 news articles, with a dependence taxonomy drawn from newsroom practice and a measurement of the copy detector itself (precision in a labelled sample, recall on edited and LLM-reworded copies). (2) \emph{Judges}: a controlled experiment with four LLM judges on their sensitivity to repeated reports, outlet diversity and supplied provenance information. (3) \emph{Continual learning}: a simulated stream with three attacks in which showing the judge one report per detected chain reduces the lies accepted from copy networks, validated on 15,084 real articles carrying inserted claims with the real detectors and two local judges. (4) \emph{Content-based dependence signals}: an entity-novelty test that recognises most copies reworded by three LLMs, compared with text and embedding similarity against other reports of the same event and an adaptive rewriter; it recognises few natural same-source derivatives. Collusion remains unsolved. The source corpus is public; code, prompts, annotations and results are available at \href{https://github.com/Metalesaek/copies-experiment}{github.com/Metalesaek/copies-experiment}; article texts are not redistributed and are rebuilt from CC-News by the scripts.

\section{Related work}\label{sec:related}

\textbf{Repetition and belief in LLMs.} \citet{churina2025layer} continually pre-train Qwen 2.5 models on corpora with increasing shares of counterfactual claims and find that belief flips rise sharply with the poison ratio; they do not compare copies with independent sources and test no defence. \citet{schuster2026whose} show that, in context, models prefer institutional sources and that repeating information from a low-credibility source can reverse that preference; their mitigation changes the model's weights. We study the same weakness at the point where a learner decides what to learn, and test a defence that changes the evidence, not the model.

\textbf{Continual learning from claims.} Our earlier paper \citep{metales2026surprise} gates each learning update with surprise, a judge, a regression test and the model's prior, and finds that the learner inherits its judge's errors; it proposes corroboration across independent chains as future work. This paper tests that proposal.

\textbf{News reuse and source criticism.} Wire copy, syndication and press-release journalism (``churnalism'') are well documented in journalism studies; newsroom verification norms such as the two-source rule count independent sources, not publications. WCEP \citep{gholipour2020wcep} collects many articles per news event for summarisation. \citet{schlichtkrull2024media} gives models background on outlets for source-critical reasoning. We use the independence of reports, rather than the reputation of outlets, as the signal.

\textbf{Text reuse in journalism.} The METER corpus measured how British newspapers reuse Press Association copy, from verbatim reprints to heavy rewriting \citep{clough2002meter}. In a study of UK national newspapers, most press articles relied wholly or mainly on agency copy or public-relations material \citep{lewis2008compromised}, a practice \citet{davies2008flat} called churnalism. Our detector uses standard near-duplicate detection with MinHash \citep{broder1997resemblance}. Work on knowledge conflicts in LLMs \citep{xu2024knowledge} studies how models resolve disagreement between context and memory; we study how they count agreeing context.

\textbf{How the task differs from deduplication.} News deduplication and event clustering group articles that are similar or cover the same event. We ask a different question of two reports of one event: does the later one add information, or does it repeat what the earlier one already contained? The answer matters when reports serve as evidence for an LLM judge or as training input for a continual learner, where grouping by event would merge independent confirmations with copies.

\textbf{Similarity, deduplication and verification.} Sentence embeddings \citep{reimers2019sbert} are the obvious alternative to our entity-and-quotation score for linking rewrites; Section~\ref{sec:novelty} compares against them, including current embedding models and a reranker, and finds that they rate independent reports of the same event nearly as similar to the original as rewrites. Deduplicating pre-training data \citep{lee2022dedup} removes near-identical text to reduce memorisation; we use duplication instead as evidence about how many sources support a claim. Claim-verification benchmarks such as FEVER \citep{thorne2018fever} separate retrieving evidence from judging a claim; our judges are given the evidence, and we study how its dependence structure is counted.

\section{How news is copied}\label{sec:copying}

Our detector links about one in five sampled news articles to a near-copy from another outlet, and most copies reflect wire services, shared ownership or syndication rather than independent reporting.

\textbf{Data and detector.} We drew a uniform random sample of 200,000 English articles longer than 300 characters, from 6,877 outlets, in CC-News (2017-2019, with outlet domain, date and URL). Each article's first 2,000 characters were reduced to 5-word shingles and hashed with MinHash (64 permutations); locality-sensitive hashing found pairs from different outlets with estimated Jaccard similarity of at least 0.6. Pairs were joined into chains (connected groups of near-copies).

\textbf{How much is copied.} The detector links 37,705 articles (18.9\%) to a near-copy from another outlet; this estimates near-duplicate text across outlets, not every form of journalistic dependence. The articles fall in 12,243 chains (connected groups of LSH candidate pairs; these chains supply the chain sizes of the simulation in Section~\ref{sec:stream}). This is a lower bound: the sample holds about 28\% of the corpus, so many copies' partners are missing. Which part of the article is compared matters little (boilerplate removed): 18.5\% of articles are linked with the first 1,000 characters, 18.9\% with the first 2,000, 19.2\% with the first 4,000, 19.4\% with the full text, and 20.0\% when an article counts as linked if its first or its last 2,000 characters match, which catches wire copy placed below a locally written opening; 98\% of the articles linked with the first 2,000 characters are also linked with the full text. In 12 sampled extra pairs per window, the extra full-text links are mostly the same story; the extra links from the last 2,000 characters include shared closing boilerplate (such as rating-agency disclaimers), so reading the tail adds false links as well as true ones (data/window\_new\_pairs.csv).

\begin{table}[htbp]\centering\small
\begin{tabular}{lc}
\toprule
Outlets per chain & Chains \\
\midrule
2 & 7,550 \\
3-5 & 3,858 \\
6-10 & 670 \\
11-50 & 165 \\
\bottomrule
\end{tabular}
\end{table}

The earliest-published articles in the detected chains most often appear on wire-service sites (Reuters) or on client sites carrying AP and Canadian Press material.

\textbf{Kinds of dependence.} We labelled 150 detected pairs and 150 control pairs (same day, different outlets, a shared title word, below the threshold) with a taxonomy drawn from newsroom practice:

\begin{table}[htbp]\centering\small
\begin{tabularx}{\linewidth}{lYc}
\toprule
Dependence & Definition & Detected pairs \\
\midrule
Wire copy & Same news-agency story (AP, Reuters, Canadian Press, PA) & 52 \\
Same owner & Titles of one media group share their own content & 40 \\
Syndication & Licensed non-agency content (public-radio network, CNN affiliates, columns) & 22 \\
Press release & Same release or statement reproduced & 8 \\
Aggregator & Another outlet's article republished & 1 \\
Detector error & Shared boilerplate (login pages, paywalls, report templates); different stories & 27 \\
\bottomrule
\end{tabularx}
\end{table}

Each pair received one label, although the kinds can overlap (a group's papers may all carry one agency story). The proportions describe this labelled sample; they are descriptive estimates, not exhaustive ground truth for all detected pairs, because the labels were drafted with an AI assistant and checked blind by two authors on a subset (end of this section). By the AI-assisted draft labels, 123 of the 150 detected pairs (82\%) reflect actual dependence; all 27 false matches in the labelled sample involve shared boilerplate. Removing, before hashing, every line that occurs in at least 20 articles of the corpus raises precision in the labelled sample to 0.97 (116 of the 119 pairs still detected), at a cost in recall of 7 real dependences no longer detected; the copy statistics hardly change (18.9\% of articles, 12,323 LSH candidate chains; removing shared lines also removes some links, which splits a few chains). By the same draft labels, 142 of the control pairs were unrelated and 8 reported the same event independently. Only wire copies are what a casual reader would call ``the same story''; same-owner sharing and syndication are as common together and are equally not independent.

\textbf{Recall.} We took 200 articles of at least 1,500 characters (boilerplate removed) and made two kinds of copy: a reprint edited the way client papers edit wire copy (dateline and agency credit removed, first line dropped, last 30\% cut), and a rewrite generated by phi4 (``in your own words, the way a different newspaper would report the same story; keep every fact, name and number''). The detector links 92\% of the edited reprints to their original (median Jaccard 0.81) and none of the rewrites (median Jaccard 0.04; the rewrites keep 75\% of the original length). The text-similarity detector catches edited reprints, but the tested rewrites generated by phi4 evaded it completely, and so do rewrites of the same 200 articles by Llama 3.2 and Gemini 2.5 Flash with the same prompt (0 of 200 each, median Jaccard 0.06).

\textbf{Chains.} Joining pairs into connected components could merge articles that are not copies of each other (A like B, B like C, A unlike C). If every candidate pair must also reach an estimated Jaccard of 0.6 (LSH alone admits some pairs just below it), there are 12,107 chains, 7,133 of them pairs. In the 4,974 chains of three or more articles the median mean pairwise similarity is 0.95; 5\% contain a pair below 0.6, joined only through other articles, and none a pair below 0.3; 98\% of their articles are a direct near-copy of the chain's earliest article. The share of articles in chains moves little with the threshold: 19.9\%, 18.6\% and 17.4\% at 0.5, 0.6 and 0.7 (18.9\% with the LSH candidates used above), with or without boilerplate removal. The threshold 0.6 was set before annotation and not tuned. A chain remains a cluster of near-duplicates, not a reconstructed lineage: two articles that each summarise one press release and a third that copies one of them form a single chain.

The dependence counts in this section use labels drafted with an AI assistant from titles and the first 1,500 characters of each article; the assistant also saw the detector's similarity score. Control pairs are drawn without repetition among articles with a known publication date. As a check, the first two authors (who are not independent of the study) then labelled 100 pairs (70 detected copies, 30 controls) without seeing the drafts, the detector scores or each other's labels. They agreed on 84 of 100 labels (Cohen's $\kappa$ = 0.80 over the nine labels) and on all 100 decisions whether a pair is a real dependence or not ($\kappa$ = 1.00); agreement with the AI drafts was $\kappa$ = 0.84 and 0.92. Of the 16 disagreements, 7 lie between same-owner sharing and wire copy, when papers of one group carry an agency story; 4 between unrelated pairs and either detector errors or independent reports of the same event; 3 involve aggregation and 2 syndication. The counts above rest on the AI-assisted drafts; the blind check covers 100 of the 300 pairs.

\section{Do LLM judges count copies?}\label{sec:judges}

Under controlled conditions (claims about invented people, invented outlets), the two local judges accept far more lies as reports and outlets go from one to three, though not monotonically beyond, and ignore a note that the copies share one source; Gemini 2.5 Pro distinguishes identical from differently worded reports and follows provenance information, as a note or a table. The experiments measure sensitivity to how evidence is presented, with provenance supplied by the experimenter; they do not show that a judge can identify independent reporting.

\textbf{Setup.} We took 320 claims from the world of our earlier paper (seed 10): 80 each of known-fact lies (CounterFact counterfactuals the student knows), legitimate updates of known facts, lies about invented people and true facts about invented people. Four judges: Llama-3.2-3B-Instruct and phi4 (14B), as 4-bit Ollama builds, and Gemini 2.5 Flash and 2.5 Pro through the API; temperature 0, answers up to 64 tokens (Gemini Pro cannot switch reasoning off and used the smallest reasoning budget, 128 tokens). Each claim was shown in nine conditions: the claim alone; one, three, five or ten reports with identical text from different invented outlets; three reports with different wording; three identical reports among seven reports about other claims (as long as the ten-copy prompt); and three or ten identical reports with the note ``all of these reports are reprints of one and the same wire story; they are one source''. The judge answers True or False; in a second pass on five conditions it may also answer Unsure. Prompts are in Appendix~\ref{sec:app-prompts}. Answers with neither word count as not accepted; we report them with the Unsure answers as abstentions: 3 for Llama 3.2 (``I cannot verify the claim''), 35 for phi4 (a reasoning chain that ran past the limit, all of them with five or ten reports in the prompt) and 29 for Gemini Pro (26 of them ``Undetermined'', all with the one-source note).

\textbf{Lies about invented people accepted as true} (binary True/False answers; n = 80 per cell; truths about invented people were accepted at nearly the same rates in every cell; the three-way True/Unsure/False ablation follows below; 95\% Wilson intervals are at most $\pm$0.11 per cell, and claim-level paired tests are reported in the findings):

\begin{table}[htbp]\centering\footnotesize
\begin{tabularx}{\linewidth}{lccccZZZZ}
\toprule
Judge & 1 report & 3 copies & 5 copies & 10 copies & 3 different wordings & 3 copies among 7 others & 3 copies + note & 10 copies + note \\
\midrule
Llama 3.2 3B & 0.06 & 0.71 & 0.55 & 0.44 & 0.34 & 0.40 & 0.74 & 0.51 \\
phi4 14B & 0.21 & 0.90 & 0.78 & 0.50 & 0.97 & 0.93 & 0.82 & 0.57 \\
Gemini 2.5 Flash & 0.93 & 1.00 & 1.00 & 1.00 & 1.00 & 0.97 & 0.79 & 0.20 \\
Gemini 2.5 Pro & 0.64 & 0.76 & 0.86 & 0.85 & 1.00 & 0.99 & 0.00 & 0.00 \\
\bottomrule
\end{tabularx}
\end{table}

No judge accepted these claims without a report (at most 0.01).

\textbf{Known facts} (share judged true, claim alone / 3 copies / 3 different wordings): known-fact lies 0.07 / 0.24 / 0.25 (phi4), 0.06 / 0.31 / 0.51 (Flash), 0.06 / 0.16 / 0.31 (Pro); legitimate updates 0.05 / 0.20 / 0.28 (phi4), 0.06 / 0.39 / 0.53 (Flash), 0.06 / 0.11 / 0.41 (Pro). The judge's prior knowledge appears to weigh most, but reports still move it, and lies and true updates move together.

\textbf{With an Unsure option} (lies about invented people, True / Unsure / False): phi4 answers 0.33 / 0.68 / 0.00 for one report and 0.96 / 0.04 / 0.00 for three copies; Gemini Pro 0.11 / 0.89 / 0.00 for one report, 0.59 / 0.41 / 0.00 for three copies and 1.00 / 0.00 / 0.00 for three differently worded reports. Copies move the judges from Unsure to True, so the effect is not an artefact of forcing a binary answer.

Five findings. Tests are exact McNemar tests on the same 80 claims, Holm-adjusted over all 112 contrasts (56 for each kind of lie). (1) \textbf{Copies act as confirmations for the local judges}: three reprints raise acceptance of lies about invented people over one report for Llama 3.2 (0.06 to 0.71; 52 claims flip to True, none back; adjusted p \textless{} \(10^{-13}\)) and phi4 (0.21 to 0.90; 55 and 0; p \textless{} \(10^{-14}\)). For Gemini Flash (0.93 to 1.00, a ceiling) and Gemini Pro (0.64 to 0.76; 22 and 12) the rise is not significant. (2) \textbf{Gemini Pro responds to the wording of the reports}: it accepts three differently worded reports more often than three identical copies (1.00 against 0.76; 19 and 0; adjusted p = 0.0004); for phi4 the difference is not significant, Flash is at ceiling, and Llama 3.2 prefers the copies (0.34 against 0.71). Different wordings from different outlets are not established independent reporting, so this shows sensitivity to report diversity, not that a judge infers where the reports come from. (3) \textbf{Gemini Pro follows the provenance note most consistently}: with the note, Gemini Pro accepts none of the lies, answering False for 0.90 and giving no verdict for 0.10 with three copies, and False for all with ten; given an Unsure option, it answers Unsure for every claim with the note. Flash also responds, to 0.79 (three copies) and 0.20 (ten), but less consistently: it ignores the note on differently worded reports (0.97, probe below); phi4 (0.82) and Llama 3.2 (0.74) do not change significantly. The probes below test whether it reasons about provenance or follows the instruction it is given. (4) \textbf{Gemini Flash believes a single report} (0.93): for it, one unverified source is already enough. (5) \textbf{Without knowledge there is no truth signal}: lies and truths about invented people are accepted at nearly the same rates, so the reports are the principal evidence available to the judges.

\textbf{Probes: what in the prompt moves the judge?} Same 80 lies about invented people, three reports each, share accepted:

\begin{table}[htbp]\centering\footnotesize
\begin{tabularx}{\linewidth}{Ycccc}
\toprule
Prompt & Llama 3.2 & phi4 & Gemini Flash & Gemini Pro \\
\midrule
1 report (reference) & 0.06 & 0.21 & 0.93 & 0.64 \\
3 copies, 3 outlets & 0.71 & 0.90 & 1.00 & 0.76 \\
3 copies, one outlet & 0.00 & 0.62 & 1.00 & 0.61 \\
3 copies + false note: independent reports & 0.75 & 0.84 & 0.95 & 0.69 \\
3 copies + one-source note & 0.74 & 0.82 & 0.79 & 0.00 \\
3 copies + provenance table, no instruction & 0.42 & 0.88 & 0.49 & 0.01 \\
3 different wordings & 0.34 & 0.97 & 1.00 & 1.00 \\
3 different wordings + note: rewrites of one story & 0.57 & 0.99 & 0.97 & 0.36 \\
3 copies from Reuters, AP, BBC News & 0.69 & 0.55 & 1.00 & 0.50 \\
3 different wordings from Reuters, AP, BBC News & 0.34 & 0.95 & 1.00 & 0.97 \\
\bottomrule
\end{tabularx}
\end{table}

The provenance table lists outlet and original source (the same wire-story number) for each report. Differences named below are significant after Holm adjustment (exact McNemar on the same claims); the others are not. (a) \textbf{Outlet diversity matters as much as repetition for the local judges}: Llama 3.2 accepts the three copies from one outlet no more often than a single report (0.00), and phi4 at 0.62 (0.21 for one report), against 0.71 and 0.90 from three outlets: Llama 3.2 responds to outlet diversity alone, phi4 to both repetition and diversity, and neither identifies independent evidence. Gemini Pro is moved by neither. (b) \textbf{A false note that copies are independent does not significantly raise acceptance for any judge.} Gemini Pro accepts identical copies labelled independent at 0.69 against 1.00 for different wordings, so it reads the wording and not only the note. (c) \textbf{Gemini Pro follows notes that lower belief}, even when they may be wrong: three different wordings labelled as rewrites of one story fall from 1.00 to 0.36. With the one-source note it rejects every claim, below its acceptance of a single unlabelled report (0.64), and true claims about invented people too (0.00): it treats the note as a warning rather than counting sources. This asymmetry is safe for a gate against lies, but a provenance note can also block true news. (d) \textbf{A provenance table has a similar effect to an explicit instruction}: Gemini Pro accepts 0.01 with the table, and Flash (0.49 against 0.79) and Llama 3.2 (0.42 against 0.74) respond to it more than to the note; phi4 ignores both. (e) \textbf{Famous agency names do not raise acceptance.} For different wordings nothing changes; identical copies under Reuters, AP and BBC News are accepted less by phi4 (0.55) and Gemini Pro (0.50), as if identical text from three rival agencies were itself suspicious.

\textbf{Many identical copies.} The two local judges accept ten copies less often than they accept three (Llama 3.2: 0.71 to 0.44; phi4: 0.90 to 0.50), with five in between. Length does not explain it: three copies hidden among seven unrelated reports, a prompt as long as ten copies, are accepted at 0.93 by phi4 and 0.97 by Flash, at rates similar to three copies alone; only Llama 3.2 also drops in this control (0.40). Nor does the answer limit: with 64 answer tokens phi4 accepts 0.50 of the ten-copy lies, against 0.44 in a first run of the same prompts with 10 tokens; answers without a verdict, mostly reasoning cut off at the limit, count as not accepted, and excluding them gives 0.61 (0.58 in the 10-token run). The two local judges accept fewer claims with ten identical reports than with three; Gemini 2.5 Pro shows no such drop (0.64, 0.76, 0.86, 0.85 for 1, 3, 5, 10 copies).

\section{Learning from a simulated stream of reports}\label{sec:stream}

In a controlled simulation, counting reports or chains does not keep lies out of a continual learner. Showing an LLM judge one report per chain cuts the lies it accepts from a copy network by 0.10 to 0.24 for all four judges, at a mean cost of no more than 4.2 percentage points of accepted truths across the tested judges (largest upper 95\% confidence bound: 5.9 points). With a text-similarity detector it does nothing against reworded copies or collusion; Section~\ref{sec:novelty} addresses reworded copies, and Section~\ref{sec:real} repeats the study with the detectors run on real articles.

\textbf{Student and claims.} We reuse the setup of our earlier paper: Qwen2.5-1.5B-Instruct with one LoRA adapter (rank 16, alpha 32, no dropout, on all attention and MLP projections), trained with AdamW (learning rate 2e-4, no weight decay) for 8 steps of batch 16 on each chunk of 16 reports plus 32 claims replayed from a buffer of up to 512 accepted claims. A claim is a sentence prefix and its object (``Ta Tasken was born in the city of'' + ``Lima''); the loss is the mean cross-entropy on the object tokens plus \(\mathrm{KL}(\text{base}\,\|\,\text{student})\) on 300 anchor sentences (weight 1, batch 16). The full configuration is in the repository. Each world holds 400 true facts about invented people, 100 legitimate updates of facts the model knows, and 171 lies (30\% of the claims; half about invented people, half CounterFact counterfactuals). Three seeds (10-12); additional ten-seed robustness runs (seeds 10-19) are reported below.

\textbf{Reports.} Each true claim is reported by 1, 2 or 3 independent chains (50\%, 30\%, 20\%; varied below); the copies per chain are drawn from the CC-News chain sizes of Section~\ref{sec:copying}, so most chains are a single article. Lies follow one of three attacks: \textbf{copy} (one origin and 10 verbatim copies), \textbf{paraphrase} (the same, each copy reworded) and \textbf{collusion} (two distinct sources that coordinate on the same false claim without copying each other). For each copy, the simulated detector either links it to its chain or misses it. We use the recall measured in Section~\ref{sec:copying}: 0.92 for lightly edited copies and 0.00 for reworded ones (a first set of runs with assumed values, 0.95 and 0.3, gave the same picture). Reports arrive over 30 days in time order. The stream is simulated: the detectors are not run on generated text; their measured rates decide, report by report, whether a copy is linked to its chain, which keeps runs comparable across seeds and judges.

\textbf{Systems.} We distinguish claims accepted for training from facts the student absorbs (measured after the stream). We test these acceptance rules: accept at every report (\emph{every report}); accept once, at the first report (\emph{first report}); accept after three reports (\emph{3 reports}); accept after two chains as the detector sees them (\emph{2 chains}); accept at the first report if a simulated judge with accuracy 0.9 approves (\emph{judge}, as in our earlier paper); accept after two chains and simulated-judge approval (\emph{2 chains + judge}); or when an LLM judge (Llama 3.2, phi4, Gemini 2.5 Flash or, for the copy attack, Gemini 2.5 Pro) approves the evidence received so far, shown either as all reports (\emph{judge, all reports}) or as one report from each detected chain, the earliest (\emph{judge, one per chain}), at most 10 reports, re-judged whenever the evidence shown changes, with answers of at most 10 tokens (Appendix~\ref{sec:app-prompts}).

\textbf{Reading the results.} Three kinds of outcome are reported: the share of distinct claims a rule \emph{accepts} for training; what the student has \emph{absorbed} after the stream (facts learned, lies absorbed); and \emph{retention} of prior knowledge. Paired differences are computed from unrounded per-seed values, so they can differ in the last digit from differences of the rounded table entries. The main runs use three seeds, so their intervals have two degrees of freedom and are rough indications of uncertainty; the ten-seed runs below are the firmer evidence where available. All results are conditional on this stream construction, the detector error model and the student setup.

\textbf{Results: share of distinct claims accepted} (mean of 3 seeds; seed standard deviation at most 0.04; measured detector; rows that do not use the detector, including phi4 with all reports, come from the first run, on the same worlds and judge verdicts):

\begin{table}[htbp]\centering\footnotesize
\begin{tabularx}{\linewidth}{lccZZ}
\toprule
System & Truths & Lies: copy & Lies: paraphrase & Lies: collusion \\
\midrule
Every report & 1.00 & 1.00 & 1.00 & 1.00 \\
First report & 1.00 & 1.00 & 1.00 & 1.00 \\
3 reports & 0.25 & 1.00 & 1.00 & 0.00 \\
2 chains & 0.50 & 0.57 & 1.00 & 1.00 \\
Simulated judge (0.9) & 0.90 & 0.12 & 0.12 & 0.12 \\
2 chains + simulated judge & 0.44 & 0.08 & 0.12 & 0.12 \\
Llama 3.2, all reports & 0.28 & 0.56 & 0.64 & 0.28 \\
Llama 3.2, one per chain & 0.24 & \textbf{0.35} & 0.64 & 0.28 \\
phi4, all reports & 0.51 & 0.57 & 0.67 & 0.60 \\
phi4, one per chain & 0.50 & \textbf{0.33} & 0.67 & 0.60 \\
Gemini Flash, all reports & 0.85 & 0.75 & 0.80 & 0.69 \\
Gemini Flash, one per chain & 0.84 & \textbf{0.65} & 0.80 & 0.69 \\
Gemini Pro, all reports & 0.69 & 0.63 & - & - \\
Gemini Pro, one per chain & 0.67 & \textbf{0.47} & - & - \\
\bottomrule
\end{tabularx}
\end{table}

Under collusion there are no copies to remove, so both variants of a judge see the same evidence.

\textbf{What the student learned (copy attack, mean of 3 seeds).} Measured after the stream as in our earlier paper (top-1 for invented people, pairwise against the true value for known facts):

\begin{table}[htbp]\centering\footnotesize
\begin{tabularx}{\linewidth}{lZZZZc}
\toprule
System & New facts learned & Updates learned & Lies absorbed: invented & Lies absorbed: known facts & Retention \\
\midrule
Every report & 0.82 & 0.98 & 0.86 & 1.00 & 0.88 \\
First report & 0.82 & 0.98 & 0.83 & 0.98 & 0.88 \\
2 chains & 0.49 & 0.53 & 0.56 & 0.58 & 0.91 \\
Simulated judge (0.9) & 0.83 & 0.89 & 0.12 & 0.19 & 0.91 \\
Llama 3.2, all reports & 0.34 & 0.22 & 0.78 & 0.36 & 0.93 \\
Llama 3.2, one per chain & 0.30 & 0.18 & \textbf{0.48} & \textbf{0.24} & 0.94 \\
phi4, all reports & 0.59 & 0.17 & 0.83 & 0.27 & 0.95 \\
phi4, one per chain & 0.56 & 0.15 & \textbf{0.51} & \textbf{0.15} & 0.95 \\
Gemini Flash, all reports & 0.88 & 0.38 & 0.88 & 0.50 & 0.92 \\
Gemini Flash, one per chain & 0.85 & 0.36 & 0.87 & \textbf{0.32} & 0.93 \\
Gemini Pro, all reports & 0.75 & 0.21 & 0.87 & 0.28 & 0.94 \\
Gemini Pro, one per chain & 0.75 & 0.20 & 0.75 & \textbf{0.16} & 0.95 \\
\bottomrule
\end{tabularx}
\end{table}

\textbf{Paired differences, one report per chain minus all reports} (copy attack, same worlds and run, 95\% t-intervals over three seeds; for phi4, all reports from the first run):

\begin{table}[htbp]\centering\footnotesize
\begin{tabularx}{\linewidth}{lZZZZ}
\toprule
Judge & Copied lies accepted & Truths accepted & Lies absorbed: invented & Lies absorbed: known facts \\
\midrule
Llama 3.2 & $-$0.22 [$-$0.27, $-$0.17] & $-$0.042 [$-$0.059, $-$0.025] & $-$0.29 [$-$0.34, $-$0.25] & $-$0.13 [$-$0.30, 0.05] \\
phi4 & $-$0.24 [$-$0.27, $-$0.22] & $-$0.016 [$-$0.026, $-$0.006] & $-$0.32 [$-$0.44, $-$0.20] & $-$0.11 [$-$0.15, $-$0.08] \\
Gemini Flash & $-$0.10 [$-$0.13, $-$0.07] & $-$0.005 [$-$0.013, 0.002] & $-$0.02 [$-$0.22, 0.19] & $-$0.18 [$-$0.25, $-$0.11] \\
Gemini Pro & $-$0.16 [$-$0.23, $-$0.09] & $-$0.012 [$-$0.021, $-$0.003] & $-$0.12 [$-$0.28, 0.04] & $-$0.13 [$-$0.19, $-$0.06] \\
\bottomrule
\end{tabularx}
\end{table}

Under the paraphrase attack the text detector links no copy, so showing one report per chain changes nothing (the intervals include zero; Section~\ref{sec:novelty} tests a detector that does better). With ten seeds (10-19) for Llama 3.2 and Gemini Flash (measured text detector), the paired differences match the three-seed table: copied lies accepted -0.23 {[}-0.24, -0.21{]} and -0.09 {[}-0.11, -0.08{]}, truths accepted -0.036 {[}-0.042, -0.031{]} and -0.005 {[}-0.007, -0.002{]}, absorbed lies about invented people -0.29 {[}-0.31, -0.27{]} and +0.01 {[}-0.03, 0.06{]}, about known facts -0.13 {[}-0.16, -0.10{]} and -0.16 {[}-0.19, -0.14{]}; Gemini Pro keeps three seeds. In a separate ten-seed run for phi4 (entity-novelty detector of Section~\ref{sec:novelty}), the paired differences match the three-seed runs of that configuration and their intervals are narrower: copied lies accepted -0.24 {[}-0.25, -0.22{]}, absorbed lies about invented people -0.31 {[}-0.34, -0.28{]}, truths accepted -0.027 {[}-0.033, -0.020{]}; under the paraphrase attack -0.25 {[}-0.28, -0.23{]} and -0.27 {[}-0.33, -0.22{]}. The three-seed intervals of the same configuration (seeds 10-12) contain the ten-seed means; the three-seed results with the text detector above are broadly consistent with them. Copy removal does not work by making the learner more conservative: in the ten-seed run, legitimate updates learned change by -0.006 {[}-0.022, 0.010{]} and retention by +0.007 {[}-0.002, 0.016{]} under the copy attack. Showing one report per chain also shortens the prompt; in the judge experiment, three copies padded to the length of ten with unrelated reports were accepted as often as three copies by phi4 and Flash (Section~\ref{sec:judges}), so length alone does not explain the effect for these judges, but we ran no length-matched control inside the stream.

\textbf{Share of single-source truths} (copy attack, 3 seeds, phi4, first-run detector values). Copy removal cuts accepted copied lies by the same amount whatever share of truths comes from one source: -0.27 {[}-0.29, -0.26{]} at 20\%, -0.29 at 50\% and -0.28 {[}-0.33, -0.23{]} at 80\% (phi4). Its cost in truths stays small: -0.010 {[}-0.023, 0.003{]} at 20\% and -0.026 {[}-0.041, -0.011{]} at 80\%. Counting chains without a judge pays the full price: it accepts 0.80, 0.50 and 0.22 of truths at 20\%, 50\% and 80\% single-source truths, while still admitting 0.43-0.46 of copied lies.

\textbf{Findings.} (1) Counting reports is the worst rule: three reports admit every copied lie but only a quarter of truths. (2) With the measured detector, counting chains alone admits 57\% of copied lies, close to the chance that the detector misses at least one of ten copies (\(1-0.92^{10}=0.57\)): every missed copy looks like a new source, so the attacker gains from sending more copies. Rewording and collusion defeat it completely, and it blocks every truth that comes from a single source. (3) Showing an LLM judge one report per chain reduces accepted copied lies for all four judges; for the two judges that weigh the number of reports most (Llama 3.2 and phi4), it cuts them by about 40\% and the lies about invented people the student absorbs by more than a third. (4) The benefit is smallest for Gemini Flash, which already believes a single report (Section~\ref{sec:judges}): removing copies cannot help a judge that is convinced by the first one, so its absorbed lies about invented people barely change, while lies about known facts fall from 0.50 to 0.32. (5) The LLM judges are much weaker than the simulated one: they accept a quarter (Llama 3.2), half (phi4) or 85\% (Flash) of true claims, and few legitimate updates are learned (15-38\%). (6) Repetition itself matters little once a lie is learned at all: learning every copy and learning a claim once absorb nearly the same share of lies about invented people (0.86 against 0.83; paired difference 0.027 {[}-0.017, 0.071{]}). (7) Gemini 2.5 Pro sits in between: copy removal cuts its accepted copied lies from 0.63 to 0.47 and its absorbed lies about known facts from 0.28 to 0.16. (8) The proportion of accepted distinct claims that are false (false discovery rate; 0.25 when everything is learned) falls with copy removal for every judge: Llama 3.2 from 0.41 to 0.33, phi4 0.28 to 0.19, Flash 0.23 to 0.21, Pro 0.24 to 0.19. With all reports shown, Llama 3.2 is worse than no filter at all; the simulated judge reaches 0.04 alone and 0.06 with two chains.

\section{Detecting LLM rewrites by entity novelty}\label{sec:novelty}

\textbf{Idea.} A rewrite changes the words but keeps most names, numbers and quotes; independent reporting of the same event usually introduces new ones, such as another witness, a figure or a quote. For a report that resembles an earlier one in topic, we therefore measure its entity-and-quotation novelty: the share of its named entities (spaCy, en\_core\_web\_sm) and quoted passages that do not appear in the earlier report. A low score marks a copy however it is worded. The score is a proxy for new information, not a measure of new facts: two reports with the same entities can state opposite things (``the minister confirmed'' and ``the minister denied''). A simpler rule-based extractor (capitalised names, numbers, quotes) serves as a check. The baselines below show that the entities carry the signal and the quotations add nothing, so the test is in effect an entity-novelty test; the stream runs, made before this ablation, keep the quotations.

\textbf{Definition.} For an earlier report \emph{o} and a later report \emph{r}, let E(x) be the set of distinct items extracted from the first 2,500 characters of x (boilerplate removed): spaCy entities of 16 types (PERSON, ORG, GPE, LOC, NORP, FAC, EVENT, PRODUCT, WORK\_OF\_ART, LAW, DATE, TIME, MONEY, PERCENT, QUANTITY, CARDINAL), lower-cased, with runs of spaces and commas collapsed and a leading ``the'' removed, and quoted passages of 15-300 characters between quotation marks, each represented by its first 40 lower-cased characters. Entities and quotations are pooled in one set, so repeated mentions count once. The novelty of \emph{r} relative to \emph{o} is \(N(r\mid o)=|E(r)\setminus E(o)|\,/\,|E(r)|\), and \emph{r} is flagged as a copy of \emph{o} when \(N(r\mid o)\) \textless{} 0.4. The score is directional: the later report is scored against the earlier one (in the held-out evaluation, a pair published on the same day is flagged if either direction falls below the threshold). Items match only exactly, so a shortened or differently cut quotation counts as new. A report with no extractable items gets \(N=0\) and is flagged; a report whose look-alike has none gets \(N=1\). Entity overlap, a baseline below, is 1 - N computed on entities only.

\textbf{Test.} We compare the 200 phi4 rewrites of Section~\ref{sec:copying} with comparison reports of the same event, found for 167 of the originals: another outlet, published within one day, sharing more than three extracted items with the original, and 5-gram Jaccard below 0.2. This automatic selection is noisy. Only 9 of the 167 pairs (5\%) carry the same agency or press-release credit, but a manual check of 40 random pairs found 11 independent reports of the same event, 7 from the same source (the same agency story, or one interview retold), 21 about a different event that merely shares names or dates, and 1 unclear. None of the 11 confirmed independent reports falls below the threshold of 0.4 (their novelty is 0.53-0.74; 0 of 11, 95\% Wilson interval {[}0, 0.26{]}), and 3 of the 7 same-source reports do, which is a correct merge. Restricted to the 11 confirmed independent reports, the AUC is unchanged: 0.98 against the plain rewrites (0.97 on all pairs) and 0.94 against the adaptive ones (0.94 on all pairs), so the easy different-event pairs do not inflate it; confirmed independent reports of the same event do, however, have lower novelty (median 0.62) than the full comparison set (0.84), so the margin above the threshold is smaller than the medians suggest. An adaptive attacker rewrites the same 200 articles with the instruction to add three or four plausible invented details (names, numbers, places, quotes) so that the copy reads like independent reporting.

\begin{table}[htbp]\centering\footnotesize
\begin{tabularx}{\linewidth}{lccZZ}
\toprule
Extracted items from & Rewriter & AUC & Rewrites caught (novelty below the fixed threshold 0.4) & Comparison reports flagged (automatic set) \\
\midrule
Rules & phi4 & 0.96 & 0.82 & 0.05 \\
spaCy & phi4 & 0.97 & 0.91 & 0.05 \\
Rules & phi4, adding invented details & 0.92 & 0.57 & 0.05 \\
spaCy & phi4, adding invented details & 0.94 & 0.74 & 0.05 \\
\bottomrule
\end{tabularx}
\end{table}

The median rewrite has a novelty of 0.21 (0.31 with invented details), against 0.84 for the automatically selected comparison set and 0.62 for the 11 confirmed independent reports. The false-alarm rate of 0.05 is measured on the automatic comparison set, which mixes independent reports, same-source reports and different events; on the confirmed independent reports it is 0 of 11 (95\% interval {[}0, 0.26{]}), too few to establish a rate. The threshold of 0.4 was fixed in the script before its first run and not tuned, and no separate validation set was used; recall and false alarms at thresholds 0.1-0.5 are in Appendix~\ref{sec:app-threshold}. The two extractors rank reports alike (rank correlation 0.88 and 0.91).

\textbf{Other rewriters.} To check that the test is not fitted to phi4, Llama 3.2 and Gemini 2.5 Flash rewrote the same 200 articles with the same prompt. With spaCy entities the AUC against the automatic comparison reports is 0.96 and 0.97 (phi4: 0.97), 0.81 and 0.90 of the rewrites fall below the fixed threshold of 0.4 (phi4: 0.91) while 5\% of the comparison reports are flagged, and against the 100 held-out independent pairs the AUC is 0.98 and 0.99 (phi4: 0.98). All three rewriters were asked to keep every name and number; a rewriter that drops or replaces entities is the adaptive case above.

\textbf{Held-out set.} Because 11 confirmed pairs are too few, we built a second set after the threshold was fixed: 250 pairs from the same sample chosen to be about the same event (two outlets, at most a day apart, similar titles, at least three shared named entities, 5-gram Jaccard below 0.2), labelled blind by the first two authors without detector scores. They agreed on 225 pairs (raw agreement 0.90, Cohen's $\kappa$ = 0.86 over four labels): 100 independent reports of the same event, 63 from the same source, 40 about different events, 22 unclear. The 25 pairs on which they disagreed (12 between independent and unclear, 5 between independent and different event, 4 between same source and unclear, 2 between same source and independent, 2 others) are excluded from every evaluation below, which uses only agreed labels.

\begin{table}[htbp]\centering\small
\begin{tabularx}{\linewidth}{lccZ}
\toprule
Agreed label & Pairs & Median novelty & Flagged as copy (novelty below 0.4) \\
\midrule
Independent & 100 & 0.68 & 0.01 (1) [0.00, 0.05] \\
Same source & 63 & 0.57 & 0.17 (11) [0.10, 0.29] \\
Different event & 40 & 0.68 & 0.05 (2) \\
Unclear & 22 & 0.62 & 0.00 \\
\bottomrule
\end{tabularx}
\end{table}

The test wrongly flags 1 of the 100 independent pairs, and against these 100 it separates the phi4 rewrites with AUC 0.98 (0.95 for the adaptive rewrites; 0.98 and 0.96 when same-day pairs are scored in the listed direction, as in the baselines below), as on the 11 confirmed pairs. It recognises few natural same-source derivatives, however: when a journalist rewrites a wire story or a press release, the article usually adds names and figures of its own, and only 17\% fall below the threshold. The test catches LLM rewrites of one article, not dependence in general; such reports still count as separate sources, an error in the attacker's favour, not the learner's.

\textbf{Baselines.} The same originals, rewrites and comparison reports, scored by standard similarity measures, including current embedding models and a reranker (higher = more likely a copy). Checkpoints on Hugging Face: sentence-transformers/all-MiniLM-L6-v2 and sentence-transformers/all-mpnet-base-v2 \citep{reimers2019sbert}, BAAI/bge-large-en-v1.5 and BAAI/bge-reranker-large \citep{xiao2024cpack}, thenlper/gte-large \citep{li2023gte} and cross-encoder/stsb-roberta-base; through the Gemini API, gemini-embedding-001 \citep{lee2025gemini} with task type SEMANTIC\_SIMILARITY and 3,072 dimensions, accessed on 5 November 2026. Each report enters with its first 2,500 characters, truncated to 512 tokens for the local models; the score is the cosine of the two embeddings or the cross-encoder's output. The fifth column uses the 100 held-out independent pairs as negatives. Recall at 5\% false alarms uses the threshold that flags 5\% of the automatic comparison set:

\begin{table}[htbp]\centering\footnotesize
\begin{tabularx}{\linewidth}{lZZZZZ}
\toprule
Detector & AUC, rewrites vs all comparison reports & AUC, rewrites vs 11 confirmed independent & AUC, adaptive vs confirmed & AUC, rewrites vs 100 held-out independent & Recall at 5\% false alarms (rewrites / adaptive; threshold set per detector, not the fixed 0.4) \\
\midrule
5-gram Jaccard (text detector) & 0.85 & 0.72 & 0.75 & 0.92 & 0.06 / 0.11 \\
TF-IDF cosine & 0.94 & 0.93 & 0.91 & 0.98 & 0.64 / 0.55 \\
Sentence embeddings, MiniLM-L6 & 0.91 & 0.70 & 0.72 & 0.87 & 0.50 / 0.52 \\
Sentence embeddings, MPNet & 0.91 & 0.77 & 0.77 & 0.87 & 0.46 / 0.47 \\
Embeddings, BGE-large v1.5 & 0.95 & 0.88 & 0.85 & 0.94 & 0.68 / 0.61 \\
Embeddings, GTE-large & 0.96 & 0.89 & 0.84 & 0.94 & 0.75 / 0.63 \\
Embeddings, Gemini embedding 001 & 0.95 & 0.94 & 0.88 & 0.97 & 0.61 / 0.42 \\
Cross-encoder (STS-B RoBERTa) & 0.86 & 0.80 & 0.83 & 0.81 & 0.36 / 0.43 \\
Reranker (bge-reranker-large) & 0.82 & 0.77 & 0.73 & 0.66 & 0.27 / 0.22 \\
Quotation overlap & 0.57 & 0.52 & 0.55 & 0.58 & 0.18 / 0.18 \\
Entity overlap & 0.97 & 0.98 & 0.97 & 0.99 & 0.91 / 0.78 \\
Entity + quotation novelty (ours) & 0.97 & 0.98 & 0.94 & 0.98 & 0.91 / 0.68 \\
Ours + GTE-large (rank average) & 0.98 & 0.97 & 0.93 & 0.98 & 0.90 / 0.74 \\
\bottomrule
\end{tabularx}
\end{table}

Current embedding models narrow the gap substantially. On the 100 held-out independent pairs they are competitive: the paired differences to our score are -0.00 {[}-0.02, 0.01{]} for TF-IDF and -0.01 {[}-0.02, 0.01{]} for the Gemini embedding (95\% stratified bootstrap intervals, 2,000 resamples; results/auc\_ci.md), while BGE and GTE are lower (-0.04 {[}-0.07, -0.02{]} each). Against the 11 confirmed pairs every other single score except entity overlap is lower than ours, the best embedding (Gemini) by -0.04 {[}-0.09, -0.01{]}, but with 11 negatives these AUCs are imprecise (ours 0.98 {[}0.96, 0.99{]}, TF-IDF 0.93 {[}0.87, 0.97{]}). At the 5\% false-alarm operating point defined on the automatic comparison set, the embeddings catch 0.46-0.75 of the rewrites against 0.91. Measures of meaning rate a confirmed independent report of the same event nearly as similar to the original as a rewrite (median cosine 0.84-0.94 against 0.89-0.98). Entity overlap is the best or tied-best single score on every set of LLM rewrites, needs no model, and is better than our score against the adaptive attacker (+0.03 {[}0.01, 0.04{]} against the confirmed pairs), whose invented quotes count as new information. In these comparisons the signal comes primarily from the entities; quotations add nothing. Combining our score with an embedding does not improve performance consistently: it raises adaptive recall at 5\% false alarms from 0.68 to 0.74 but changes no AUC on the LLM rewrites significantly. No tested score separates natural same-source derivatives from independent reports well (63 and 100 held-out pairs): 5-gram Jaccard reaches AUC 0.90 {[}0.85, 0.95{]} because some derivatives keep long passages, TF-IDF 0.80, embeddings and cross-encoders 0.61-0.76, and the entity tests 0.61-0.63, significantly below Jaccard, TF-IDF and the larger embeddings.

\textbf{In the simulated stream.} We rerun the learning study with phi4 as judge and the measured spaCy test as the detector: a reworded copy is missed with probability 0.09 (0.26 for the adaptive rewriter), the share of the 200 single rewrites not flagged (in the generated networks below, with other articles and sampled rewrites, and where a copy can also be linked through a copy already linked, the missed shares are 0.06 and 0.20), an independent chain is wrongly merged with another with probability 0.05, a simulation assumption that takes the flag rate on the automatic comparison set as a conservative proxy (the held-out set above gives 0.01, too few pairs for a precise rate), and verbatim copies keep the text detector's recall of 0.92.

\begin{table}[htbp]\centering\footnotesize
\begin{tabularx}{\linewidth}{YccZZZ}
\toprule
System & Truths & Lies: copy & Lies: paraphrase & Lies: paraphrase, adaptive & Lies: collusion \\
\midrule
phi4, all reports & 0.51 & 0.57 & 0.67 & 0.67 & 0.60 \\
phi4, one per chain, text detector & 0.50 & 0.33 & 0.67 & 0.67 & 0.60 \\
phi4, one per chain, entity-novelty test & 0.49 & 0.33 & \textbf{0.41} & \textbf{0.59} & 0.58 \\
2 chains, entity-novelty test & 0.48 & 0.57 & 0.64 & 0.95 & 0.94 \\
2 chains + simulated judge, entity-novelty test & 0.43 & 0.08 & 0.09 & 0.12 & 0.11 \\
\bottomrule
\end{tabularx}
\end{table}

Against all reports, one report per chain with the entity-novelty test lowers accepted paraphrased lies by 0.27 {[}0.18, 0.35{]} (0.673 to 0.407 before rounding; all reports from the same run) and the lies about invented people the student absorbs from 0.92 to 0.63 (difference 0.29 {[}0.05, 0.52{]}); the adaptive rewriter shrinks the gain to 0.08 {[}0.01, 0.15{]}. The cost in truths is 2.5 points {[}1.8, 3.3{]}. Wrong merges of independent chains cost the judge-free rules little: counting chains accepts 0.48 of truths against 0.50 with the text detector, and with the simulated judge 0.43 against 0.44. Why so little remains of a test that catches 91\% of rewrites: with ten copies the judge still sees at least two reports whenever one copy slips through, which happens with probability \(1-0.91^{10}=0.61\). Under collusion nothing changes, as expected: the two liars coordinate but do not copy each other, so the dependence signals tested here cannot merge them. Counting chains alone remains insufficient, accepting most lies under every attack. Drawing misses independently for each copy is pessimistic here: when the test misses all copies of a lie together or none, at the same rate per copy, accepted paraphrased lies fall from 0.41 to 0.16 (difference -0.25 {[}-0.38, -0.12{]}) and absorbed lies about invented people from 0.63 to 0.27, and counting chains alone accepts only 0.09, because a lie now passes only when its whole chain evades the test (0.095).

\textbf{Generated rewrite networks.} To test the assumption that detector misses are independent, we generated networks of ten phi4 rewrites for each of 20 original articles (temperature 0.8, seeds 1-10), plainly and with invented details, and both detectors processed each network in stream order, linking a copy when it matches the original or any copy already linked:

\begin{table}[htbp]\centering\footnotesize
\begin{tabularx}{\linewidth}{lccZZZ}
\toprule
Rewrites & Detector & Copies missed & Networks with no copy missed & Networks leaking at least one copy & Same, if misses were independent \\
\midrule
Plain & Text similarity & 1.00 & 0.00 & 1.00 & 1.00 \\
Plain & Entity-novelty test & 0.06 & 0.85 & 0.15 & 0.46 \\
Invented details & Text similarity & 1.00 & 0.00 & 1.00 & 1.00 \\
Invented details & Entity-novelty test & 0.20 & 0.50 & 0.50 & 0.90 \\
\bottomrule
\end{tabularx}
\end{table}

Detector misses are concentrated in a few networks, suggesting that evasion depends more on the original article than on random variation between rewrites. We reran the stream with these measured patterns, each paraphrased lie taking the miss pattern of a randomly drawn generated network (phi4 judge, 3 seeds). Twenty networks from one rewriting model are too few to characterise attackers in general. Accepted paraphrased lies fall from 0.67 with all reports shown to 0.20 (difference -0.48 {[}-0.59, -0.36{]}), against 0.41 with independent misses, and the absorbed lies about invented people from 0.92 to 0.32. With invented details the judge accepts 0.40 (-0.28 {[}-0.42, -0.14{]}; 0.59 with independent misses). The cost in truths is 2.5 points as before, and counting chains alone accepts 0.17 and 0.54. For these generated networks, independent misses were the more pessimistic case; whether this holds for other rewriting models or attackers is unknown. The stream's own reports are still single sentences: the generated networks supply the pattern of misses, not the text the judge reads.

\textbf{Sensitivity to false merges.} The stream assumes a false-merge probability of 0.05, a conservative proxy taken from the flag rate on the automatic comparison set (the held-out set suggests 0.01). Raising it to 0.10, 0.20 and 0.30 (phi4 judge, one report per chain, 3 seeds) leaves accepted copied and paraphrased lies unchanged (0.33 and 0.41), since false merges touch only independent chains, and costs truths: 0.49 at 0.05, then 0.47, 0.45 and 0.42. Under collusion, merging the two liars helps: accepted lies fall from 0.58 to 0.56, 0.52 and 0.48. Under the copy attack, the share of lies among accepted claims rises from 0.19 to 0.21 at 0.30.

\section{The pipeline on real articles carrying inserted claims}\label{sec:real}

Sections~\ref{sec:stream} and~\ref{sec:novelty} drew each detector decision at a measured rate. Here the detectors run on real news articles, and no detector decision is drawn at random. The claims remain controlled: each is inserted into the articles that carry it, so this is not an evaluation on a naturally occurring stream of news claims, and only the two local judges were run. On real articles, showing the judge one report per detected chain cut the lies phi4 accepted from 0.57 to 0.09 under the copy attack, from 0.67 to 0.14 under the paraphrase attack and from 0.67 to 0.34 when the rewriter invents details, and those Llama 3.2 accepted from 0.56-0.62 to 0.07, 0.09 and 0.25. The cost in accepted truths was 2.4 points for phi4 and 10 points for Llama 3.2. Collusion was unaffected. The copy-attack figures are partly circular, as explained below; the rewrite figures are not.

\textbf{Carriers.} Every report is a real CC-News article (boilerplate removed, first 2,500 characters) with the claim sentence inserted after its first line. Copies of a chain repeat their origin's wording of the claim; rewrites use another wording. A true claim reported by one chain is carried by a real CC-News copy chain (an origin and its reprints by other outlets) whose size is drawn from the measured distribution, or, for a chain of size one, by one of 3,000 articles that no other outlet copied. A true claim reported by two or three chains is carried by one of the 100 held-out pairs that both authors labelled independent reports of the same event (Section~\ref{sec:novelty}), each article with its real reprints if it has any; a third chain is an unrelated article. Lies are carried by a real chain with at least 11 outlets (copy attack; 151 chains), by the generated networks of ten phi4 rewrites, plain or with invented details (paraphrase and adaptive attacks; 20 networks each, Section~\ref{sec:novelty}), or by an independent same-event pair (collusion). The carriers comprise 15,084 articles; their texts are rebuilt from CC-News by the released scripts.

\textbf{Detection.} The reports of a claim are processed in time order. A report joins the chain of the first earlier report it matches, by 5-gram Jaccard of at least 0.6 (text detector) or, in the full pipeline, also by entity novelty below 0.4 (the entity-novelty test of Section~\ref{sec:novelty}, ``novelty test'' in the tables); otherwise it starts a new chain. Worlds, student, judges, prompts and metrics are those of Section~\ref{sec:stream}, with phi4 and Llama 3.2 as judges and seeds 10-12. The judge still reads the outlet and the claim sentence, not the article, so differences from Section~\ref{sec:stream} come from the detection alone.

\textbf{What the detectors did} (three seeds; share of copies linked to their chain):

\begin{table}[htbp]\centering\footnotesize
\begin{tabularx}{\linewidth}{Ycc}
\toprule
Reports & Text detector & Text detector + novelty test \\
\midrule
Lie copies, copy attack (real reprints) & - & 0.999 (5,126 of 5,130) \\
Lie copies, plain rewrites & 0.000 (0 of 5,130) & 0.989 (5,076 of 5,130) \\
Lie copies, rewrites with invented details & 0.000 (0 of 5,130) & 0.878 (4,502 of 5,130) \\
Copies of true claims (real reprints) & 0.957 (2,990 of 3,125) & 0.995 (3,109 of 3,125) \\
Independent chains of true claims wrongly merged & 0 of 1,038 & 12 of 1,038 (1.2\%) \\
Colluding chains wrongly merged & - & 5 of 513 (1.0\%) \\
\bottomrule
\end{tabularx}
\end{table}

The wrong-merge rate of 1.2\% matches the 1 in 100 measured on the held-out independent pairs. The linking of the copy-attack chains is partly circular: these chains were found by the same text detector in Section~\ref{sec:copying}, so they are linked almost by construction (its recall on independently edited reprints is 0.92). The rewrites were generated independently of the detectors, so their rows are not circular.

\textbf{Results: share of distinct claims accepted} (mean of 3 seeds; seed standard deviation at most 0.04):

\begin{table}[htbp]\centering\footnotesize
\begin{tabular}{lccccc}
\toprule
System & Truths & \multicolumn{4}{c}{Lies accepted, by attack} \\
\cmidrule(lr){3-6}
 & & Copy & Paraphrase & Adaptive & Collusion \\
\midrule
phi4, all reports & 0.51 & 0.57 & 0.67 & 0.67 & 0.62 \\
phi4, one per chain (text detector) & 0.49 & - & 0.67 & 0.67 & - \\
phi4, one per chain (+ novelty test) & 0.49 & \textbf{0.09} & \textbf{0.14} & \textbf{0.34} & 0.61 \\
Llama 3.2, all reports & 0.34 & 0.56 & 0.62 & 0.62 & 0.30 \\
Llama 3.2, one per chain (+ novelty test) & 0.24 & \textbf{0.07} & \textbf{0.09} & \textbf{0.25} & 0.30 \\
2 chains (+ novelty test) & 0.50 & 0.01 & 0.10 & 0.45 & 0.99 \\
2 chains + simulated judge (+ novelty test) & 0.44 & 0.00 & 0.02 & 0.05 & 0.12 \\
\bottomrule
\end{tabular}
\end{table}

\textbf{Paired differences, one report per chain minus all reports} (entity-novelty test, same worlds, 95\% t-intervals over three seeds, from unrounded per-seed values). Truths accepted change by -0.024 {[}-0.046, -0.002{]} for phi4 and -0.104 {[}-0.174, -0.034{]} for Llama 3.2 (the true claims are the same under every attack):

\begin{table}[htbp]\centering\footnotesize
\begin{tabularx}{\linewidth}{llZZZ}
\toprule
Judge & Attack & Lies accepted & Lies absorbed: invented & Lies absorbed: known facts \\
\midrule
phi4 & copy & -0.47 [-0.53, -0.41] & -0.66 [-0.75, -0.57] & -0.15 [-0.19, -0.10] \\
phi4 & paraphrase & -0.52 [-0.61, -0.44] & -0.67 [-0.73, -0.61] & -0.25 [-0.39, -0.11] \\
phi4 & adaptive & -0.33 [-0.34, -0.31] & -0.43 [-0.47, -0.38] & -0.14 [-0.24, -0.04] \\
phi4 & collusion & -0.01 [-0.02, 0.01] & 0.00 [-0.12, 0.12] & 0.00 [-0.01, 0.02] \\
Llama 3.2 & copy & -0.50 [-0.61, -0.39] & -0.66 [-0.75, -0.57] & -0.24 [-0.37, -0.10] \\
Llama 3.2 & paraphrase & -0.53 [-0.62, -0.44] & -0.65 [-0.73, -0.58] & -0.30 [-0.41, -0.19] \\
Llama 3.2 & adaptive & -0.37 [-0.46, -0.29] & -0.48 [-0.55, -0.41] & -0.22 [-0.31, -0.13] \\
Llama 3.2 & collusion & 0.00 [0.00, 0.00] & 0.00 [-0.08, 0.09] & 0.01 [-0.07, 0.08] \\
\bottomrule
\end{tabularx}
\end{table}

\textbf{Findings.} (1) With all reports shown, the real articles reproduce the simulated stream (phi4 accepts 0.57, 0.67 and 0.62 of copied, paraphrased and colluding lies, against 0.57, 0.67 and 0.60 in Section~\ref{sec:stream}). (2) With one report per chain, the gains are larger than in the simulation. With independent misses, the simulation gave 0.33, 0.41 and 0.59 for copies, rewrites and rewrites with invented details; with the miss patterns of the generated networks, it gave 0.20 and 0.40 for the two kinds of rewrite. On real articles the values are 0.09, 0.14 and 0.34. On real articles a copy is also linked through any copy already linked, and misses concentrate in a few networks, so fewer lies reach the judge as two sources. For the copy attack, part of the gain comes from the circular linking noted above. The lies about invented people that the student absorbs fall from 0.82 to 0.16 (copy), 0.93 to 0.26 (paraphrase) and 0.93 to 0.50 (adaptive) for phi4, and from 0.76 to 0.10, 0.81 to 0.16 and 0.81 to 0.33 for Llama 3.2. The share of accepted claims that are false falls from 0.27 to 0.06 (phi4, copy attack) and from 0.36 to 0.09 (Llama 3.2). (3) The entity-novelty test is what makes this possible: the text detector alone links none of the 10,260 reworded copies, and the judge then sees exactly what it sees with all reports. Against the text detector alone, the test lowers accepted paraphrased lies by 0.52 {[}0.44, 0.61{]} and adaptive ones by 0.33 {[}0.31, 0.34{]}, at a cost of 0.4 points of accepted truths {[}-0.1, 0.9{]}. (4) The cost in truths depends on the judge. Llama 3.2, which weighs the number of reports most (Section~\ref{sec:judges}), loses copies of true claims too and accepts 10 points fewer truths; under collusion, where no lie copies are removed, its share of lies among accepted claims rises from 0.23 to 0.30. For phi4 the cost is 2.4 points, legitimate updates learned change by -0.02 {[}-0.04, -0.01{]} (copy attack), retention does not fall, and accepted truths reach the student on average 0.1 days later. (5) Counting chains without a judge remains unreliable: it admits 0.45 of adaptive lies and nearly every colluding one. (6) Collusion is unaffected, as in the simulation.

\textbf{Remaining limits of this run.} The articles are real, but the claim is inserted into them, so every report of a claim shares the claim's names; this makes copies and independent reports more alike, although wrong merges stayed at 1.2\%. The 20 rewrite networks are reused across the 171 lies of each world. The judge reads the claim sentence, not the article. A stream of naturally occurring claims, with copying as it occurs in the news, remains to be tested.

\section{Discussion}\label{sec:discussion}

\textbf{The effect of provenance information varies across judges.} Llama 3.2 and phi4 count reports and ignore a note that the reports share a source, so for them a note in the prompt is no defence; changing the evidence they see is. Gemini 2.5 Pro distinguishes identical from differently worded reports and follows provenance metadata, as a note or as a plain table, but only in the direction of doubt: it ignores a false claim of independence and obeys a note of shared origin even below its trust in a single report. With Gemini 2.5 Pro, provenance in the prompt is therefore useful if it is correct; a wrong one-source note blocks true news as well. For Llama 3.2 and Gemini Flash a provenance table works better than an instruction (phi4 ignores both), and the two local judges respond to outlet diversity as well as repetition, without reliably identifying whether the reports provide independent evidence. Copy removal reduced accepted copy attacks for all four judges and fits where learners sit in practice: between a news feed and a model, where outlet, timing and text are available.

\textbf{Lexical similarity is insufficient; entity overlap adds a signal.} A text-similarity detector never recognised an article rewritten by a language model. Asking whether a report adds new entities and quotations recognised most phi4 rewrites and cut the reworded-copy lies the judge accepted from 0.67 to 0.41 with independent misses, to 0.20 with the miss patterns of the generated networks and to 0.14 on real articles, but an attacker who invents details weakens it, it recognised only 17\% of natural same-source derivatives, so it detects the tested LLM rewrites and not information dependence in general, and when misses are independent, one missed copy among ten is enough to show the judge two sources. Colluding sources are distinct and coordinated by construction; because they do not copy one another, they evade the dependence signals evaluated here. Stronger defences need evidence beyond the article: agency credits and datelines, ownership of outlets, publication timing (which may reveal copies that follow their origin within hours), and shared errors or quotes. These are the signals journalists use, and our taxonomy names where each applies: wire copies carry an agency credit, same-owner sharing follows the ownership register, press releases share a source document.

\textbf{Dependence detection does not replace reliable claim verification.} Copy removal is a layer that filters evidence before a judge sees it; it is not a fact-checker. The local judges accepted only a quarter to half of true claims and few legitimate updates, consistent with our earlier finding that judges defend their own beliefs; Gemini Flash accepted most truths but also believed a single unverified report, which no amount of copy removal can fix. For claims about unknown entities no judge had a truth signal of its own. Given an Unsure option, phi4 and Gemini Pro used it for a single report, and copies moved them from Unsure to True. A continual learner fed by news needs a judge that can say ``not enough independent evidence yet'' and wait, and a pipeline that counts the evidence correctly before asking. Copy removal fixes one failure, counting copies as sources; it does not make an unreliable judge reliable, which is why Gemini Flash, already convinced by one report, gains least. The reverse holds for a judge that counts reports: on real articles Llama 3.2 gained most against copied lies but also lost 10 points of accepted truths, because removing copies removes the repetition that persuaded it of true claims as well.

\textbf{Counting rules alone fail in both directions.} A threshold on reports is beaten by copies; a threshold on chains is beaten by the detector's own misses, even with the entity-novelty test, and it blocks true news that only one outlet carries. Independence is useful as an input to judgement, not as a substitute for it, and it is not truth: ten independent reports can repeat the same error, and one report can be right.

\section{Limitations}\label{sec:limitations}

\begin{itemize}
\tightlist
\item
  \textbf{Detector in the learning study.} Sections~\ref{sec:stream} and~\ref{sec:novelty} enter measured detector rates as per-report probabilities. Section~\ref{sec:real} runs the detectors on real articles, but the claim sentence is inserted into them, 20 rewrite networks are reused across lies, the copy-attack chains were found by the text detector itself, and only phi4 and Llama 3.2 judge. Recall on reworded copies was measured with three rewriting models (phi4, Llama 3.2, Gemini 2.5 Flash) and one prompt, and with one adaptive prompt (phi4); the stream uses the phi4 rates, and a stronger attacker could invent more. The false-alarm rate of the entity-novelty test on independent reports rests on 100 pairs labelled by the first two authors, who are not independent of the study, and the test recognises only 17\% of natural same-source derivatives. The entity-novelty test was run in the simulated stream with phi4 only, and on real articles with phi4 and Llama 3.2. Its score counts entities and quotations, not propositions, so it cannot tell a report that contradicts its look-alike from one that repeats it. The detectors were run on 20 generated networks of ten rewrites, whose miss patterns drive the simulated stream; in Section~\ref{sec:real} the same networks carry the lies.
\item
  \textbf{Seeds and detector errors.} The main learning results use three seeds (10-12), so their paired t-intervals have two degrees of freedom. Additional ten-seed robustness runs are reported for Llama 3.2 and Gemini Flash with the text-similarity detector and for phi4 with the entity-novelty detector; Gemini 2.5 Pro and the remaining settings use three seeds. Detector misses are drawn independently for each copy unless stated; correlated misses and the miss patterns of the generated rewrite networks made the defences stronger, not weaker (Section~\ref{sec:novelty}), which need not hold for every attacker.
\item
  \textbf{Share of single-source truths.} We assume half of true claims come from one chain and varied it (20\%, 50\%, 80\%) for phi4 only.
\item
  \textbf{Invented claims and outlets.} Claims about invented people and invented outlet names isolate the effect of report counts, but real articles carry several claims, hedges and context, and real readers and judges also weigh outlet reputation; the acceptance rates should not be read as those of a deployed fact-checker. The local judges are 4-bit builds and the Gemini judges API models, so model family, quantisation and serving are confounded.
\item
  \textbf{Models.} One student (1.5B) and four LLM judges; the strongest judge in our evaluation (Gemini 2.5 Pro) was used in the learning stream for the copy attack only. Three seeds per setting, except the ten-seed runs above.
\item
  \textbf{Answer length.} The learning runs with local judges allowed 10 answer tokens, which cut some of phi4's answers to many reports (counted as rejections); this can only understate how often phi4 accepts copied lies with all reports shown.
\item
  \textbf{Annotation.} Dependence labels were drafted with an AI assistant; the first two authors labelled 100 pairs blind ($\kappa$ = 0.80) and the 250 held-out pairs of Section~\ref{sec:novelty} ($\kappa$ = 0.86), but the authors are not independent of the study, and the boundary between same-owner sharing and wire copy is not sharp.
\end{itemize}

\section{Conclusion}\label{sec:conclusion}

Our detector links about one sampled news article in five to a near-copy from another outlet. The two local LLM judges treat such copies as confirmation: three reprints of a claim about an unknown person raised acceptance of lies from 0.21 to 0.90 (phi4), and a note that the reports share one source did not undo it. Gemini 2.5 Pro was the most responsive to provenance information and distinguished identical from differently worded reports more consistently than the other judges; Gemini Flash believes a single report. In a simulated continual learner, removing copies before the judge sees the evidence reduced the lies accepted from a copy network across all tested judges and seeds, most for the judges that count reports, while counting reports or chains without a judge was unreliable. Rewrites by phi4, Llama 3.2 and Gemini Flash evade the tested text-similarity detector; an entity-novelty test recognises most of them, on the tested sets as well as TF-IDF and the best tested embedding model, but only 17\% of natural same-source derivatives, and with the miss patterns of generated rewrite networks cut the reworded-copy lies accepted from 0.67 to 0.20 (0.40 when the rewriter invents details). Colluding sources still pass. The next step is to recognise dependence the way journalists do, from credits, ownership and timing as well as content, and to give judges the option to wait for independent evidence. Run on real news articles with the real detectors, the pipeline kept out most copied and reworded lies for both local judges, at a cost in accepted truths of 2.4 points for phi4 and 10 points for Llama 3.2; the claims in that run were inserted into the articles, so a stream of naturally occurring claims remains to be tested.

\section*{Declaration of generative AI and AI-assisted technologies in the manuscript preparation process}

During the preparation of this work the authors used Claude (Anthropic) to help draft and edit the manuscript text, to write and debug the experiment code, and to suggest additional experimental checks. The research question, the choice of experiments and the conclusions are the authors'. The authors reviewed and edited all text, checked the reported results against the released code and saved outputs, verified the references, and take full responsibility for the content of the publication.

\bibliographystyle{plainnat}
\bibliography{references}

\appendix

\section{Judge prompts}\label{sec:app-prompts}

System message (judge experiment and LLM judges in Section~\ref{sec:stream}):

\begin{verbatim}
You are a careful fact-checker. You may be shown news reports. Decide whether the claim is true. Answer with one word: True or False.
\end{verbatim}

User message, condition ``3 copies + note'' (the note line is absent in other conditions; ``none'' shows only the claim and the question):

\begin{verbatim}
News reports:
1. [Metro Tribune] Ta Tasken was born in the city of Lima.
2. [Riverside News] Ta Tasken was born in the city of Lima.
3. [Daily Courier] Ta Tasken was born in the city of Lima.
Note: all of these reports are reprints of one and the same wire story; they are one source, not 3.

Claim: Ta Tasken was born in the city of Lima.
Is the claim true?
\end{verbatim}

The note is part of the user message, directly after the reports. In ``3 different wordings'' (``indep3'' in the code) the three reports use three different wordings of the claim; in ``3 copies among 7 others'' the seven other reports state claims about other subjects, in random order.

With the Unsure option the system message ends: ``Answer with one word: True, False, or Unsure if the evidence is not enough to decide.''

\textbf{Decoding and models.} Temperature 0. The judge experiment (Section~\ref{sec:judges}) allowed 64 answer tokens; Gemini 2.5 Pro cannot switch reasoning off and used its smallest reasoning budget (128 tokens) before the answer. The LLM judges in the learning stream (Section~\ref{sec:stream}) allowed 10 answer tokens, the setting of our first judge runs. The verdict is the first standalone ``true'' or ``false'' in the answer (with the Unsure option: the first of ``true'', ``false'', ``unsure''); answers with none count as not accepted and are reported as abstentions. Models: Ollama builds phi4 (14B) and llama3.2 (Llama-3.2-3B-Instruct), both 4-bit; Gemini API model ids gemini-2.5-flash and gemini-2.5-pro, queried between 31 October and 1 November 2026.

\section{Per-seed results}\label{sec:app-seeds}

Share of copied lies accepted, all reports / one report per chain, per seed (copy attack unless stated). Detectors differ by column: Llama 3.2, Gemini Flash and Gemini Pro use the measured text-similarity detector; both phi4 columns come from the ten-seed robustness run with the entity-novelty detector of Section~\ref{sec:novelty} (the three-seed phi4 results in the Section~\ref{sec:stream} tables use the text-similarity detector):

\begin{table}[htbp]\centering\footnotesize
\begin{tabularx}{\linewidth}{lccccZ}
\toprule
Seed & Llama 3.2 & phi4 & Gemini Flash & Gemini Pro & phi4, paraphrase attack \\
\midrule
10 & 0.57 / 0.37 & 0.58 / 0.33 & 0.75 / 0.63 & 0.63 / 0.49 & 0.67 / 0.41 \\
11 & 0.54 / 0.33 & 0.57 / 0.33 & 0.74 / 0.65 & 0.61 / 0.47 & 0.65 / 0.42 \\
12 & 0.58 / 0.34 & 0.57 / 0.33 & 0.75 / 0.65 & 0.64 / 0.46 & 0.70 / 0.40 \\
13 & 0.56 / 0.35 & 0.57 / 0.34 & 0.74 / 0.67 & - & 0.67 / 0.44 \\
14 & 0.64 / 0.42 & 0.63 / 0.39 & 0.77 / 0.68 & - & 0.74 / 0.49 \\
15 & 0.56 / 0.29 & 0.52 / 0.28 & 0.72 / 0.63 & - & 0.60 / 0.37 \\
16 & 0.60 / 0.37 & 0.56 / 0.36 & 0.75 / 0.65 & - & 0.65 / 0.45 \\
17 & 0.54 / 0.30 & 0.54 / 0.29 & 0.71 / 0.64 & - & 0.68 / 0.37 \\
18 & 0.57 / 0.35 & 0.56 / 0.30 & 0.71 / 0.64 & - & 0.65 / 0.42 \\
19 & 0.54 / 0.34 & 0.58 / 0.36 & 0.76 / 0.63 & - & 0.70 / 0.44 \\
\bottomrule
\end{tabularx}
\end{table}

One report per chain accepts fewer copied lies in every seed for every judge. Each seed draws a new world (claims, chain sizes, report order) and a new training order; the claim-level judge experiment (Section~\ref{sec:judges}) is reported separately with tests on the same 80 claims. All per-run metrics are in the repository (results/*/results.jsonl).

\section{Threshold of the entity-novelty test}\label{sec:app-threshold}

Share flagged as a copy when novelty falls below each threshold (spaCy entities and quotations). The threshold 0.4 used throughout was fixed before the first run. Rewrites: 200 articles rewritten by phi4, plainly and with invented details. Automatic comparison set: 167 pairs. Held-out pairs: agreed labels of Section~\ref{sec:novelty}.

\begin{table}[htbp]\centering\footnotesize
\begin{tabular}{cccccc}
\toprule
Threshold & Plain & Adaptive & Automatic & Held-out & Held-out \\
 & rewrites caught & rewrites caught & set flagged & independent flagged & same source flagged \\
\midrule
0.1 & 0.12 & 0.03 & 0.00 & 0 of 100 & 0.00 \\
0.2 & 0.45 & 0.15 & 0.01 & 0 of 100 & 0.06 \\
0.3 & 0.76 & 0.47 & 0.03 & 0 of 100 & 0.11 \\
0.4 & 0.91 & 0.74 & 0.05 & 1 of 100 & 0.17 \\
0.5 & 0.95 & 0.86 & 0.07 & 5 of 100 & 0.30 \\
\bottomrule
\end{tabular}
\end{table}

\end{document}